\pdfoutput=1

\documentclass[11pt]{article}

\usepackage{acl}
\usepackage{booktabs}
\usepackage{multirow}
\usepackage{comment}
\usepackage{times}
\usepackage{latexsym}
\usepackage{tcolorbox}
\usepackage{caption}
\usepackage{lipsum} 

\usepackage{tcolorbox}
\usepackage{xcolor}
\usepackage{fontawesome5}
\usepackage{hyperref}

\definecolor{myred}{RGB}{200,0,0}
\definecolor{mygreen}{RGB}{0,120,0}
\usepackage[T1]{fontenc}

\usepackage[utf8]{inputenc}

\usepackage{microtype}
\usepackage{hyperref}
\usepackage{tablefootnote}
\usepackage{xspace}
\usepackage[normalem]{ulem}
\usepackage{inconsolata}

\usepackage{graphicx}
\usepackage{amsmath}

\title{An Analysis of Large Language Models for \\Simulating User Responses in Surveys}

\newcommand{\ours}{\textsc{ClaimSim}\xspace}
\newcommand{\chainot}{\textsc{CoT}\xspace}
\newcommand{\standalone}{\textsc{Direct Prompting}\xspace}

\author{
  \textbf{Ziyun Yu\textsuperscript{12}}
  \textbf{Yiru Zhou\textsuperscript{1}}
  \textbf{Chen Zhao\textsuperscript{12}}
  \textbf{Hongyi Wen\textsuperscript{12}} \\[0.5ex]
  \textsuperscript{1}Center for Data Science, NYU Shanghai \textsuperscript{2}New York University\\[0.5ex]
}

\begin{document}
\maketitle
\footnote{Corresponding author: Hongyi Wen <hw3242@nyu.edu>}

\begin{abstract}

Using Large Language Models (LLMs) to simulate user opinions has received growing attention. Yet LLMs, especially trained with reinforcement learning from human feedback (RLHF),  are known to exhibit biases toward dominant viewpoints, raising concerns about their ability to represent users from diverse demographic and cultural backgrounds. In this work, we examine the extent to which LLMs can simulate human responses to cross-domain survey questions through direct prompting and chain-of-thought (\chainot) prompting. We further propose a claim diversification method (\ours), which elicits viewpoints from LLM parametric knowledge as contextual input. Experiments on the survey question answering task indicate that, while \ours produces more diverse responses, both approaches struggle to accurately simulate users. Further analysis reveals two key limitations: (1) LLMs tend to maintain fixed viewpoints across varying demographic features, and generate single-perspective claims; and (2) when presented with conflicting claims, LLMs struggle to reason over nuanced differences among demographic features, limiting their ability to adapt responses to specific user profiles.\footnote{Our code and data are available at \url{https://github.com/Ziyun-Yu/ClaimSim}.}



\end{abstract}







\section{Introduction}

\begin{figure}[!t]
    \centering
    \includegraphics[width=1.0\linewidth ,trim={2cm 10cm 1cm 10cm},clip]{Graphics/claimsim_application.pdf}
    \caption{Top: A survey question answering example where LLMs are instructed to simulate individual user responses over diverse demographic profiles. Middle: We study two LLM-based approaches on this task, \chainot and \ours. Bottom: \ours produces more diverse answers, while both approaches struggle to simulate users accurately (slightly above random).  }
    \label{fig:main_fig}
    \vspace{-0.5cm}
\end{figure}

The development of large language models (LLMs) has enabled simulating human behavior and replicating individual decision-making processes~\cite{park2023generative, binz2024turning, aher2023using}. For example, LLMs have been adopted to design large-scale market surveys and attempt to simulate responses across diverse demographic groups, providing a cost-effective alternative to traditional survey methods~\cite{brand2024using}.



Despite their strong potential, LLMs, particularly those trained with reinforcement learning from human feedback (RLHF)~\citep{rlhf}, are known to exhibit significant bias issues~\cite{schramowski2022value, messeri2024artificial, hu2024generative}. This is especially pronounced when interacting with underrepresented demographic or cultural groups~\cite{wang2025large, Santurkar2023bias2}. For instance, \citet{wang2024not} argues that LLM responses across different languages tend to align more closely with English-speaking cultural norms.

In this study, we systematically evaluate to what extent LLMs simulate human responses to survey questions, particularly in datasets like the World Values Survey \cite{wvs2020} where respondents exhibit diverse and complex demographic profiles. As an example in Figure~\ref{fig:main_fig}, LLMs are instructed to simulate a Christian male user with a lower level of education, and to answer a gender related survey question.



We compare three categories of LLM-based approaches. First, we use direct prompting~\cite{beck2024sensitivity} as a baseline to evaluate the models’ ability to do user simulations. Then, we incorporate chain-of-thought prompting (CoT)~\cite{wei2022chain} to elicit LLMs' capability of reasoning over user demographic features and simulating the answers accordingly. We further hypothesize that eliciting multiple viewpoints from LLM parametric knowledge as context leads to more comprehensive reasoning over diverse demographic features. Therefore, we present a two-step pipeline (\ours): \textbf{(1) Claims Generator:} for each demographic feature, the LLM generates diverse claims and summarizes them, with instructions to explicitly highlight consistencies and contradictions; \textbf{(2) Answer Generator:} the summaries across all demographic features are used as additional context for the final answer simulation.

We experiment on randomly selected individuals and their survey answers drawn from the three domains: gender, politics, and religion. The results indicate that compared with \standalone and \chainot, \ours provides more diverse answer distributions. By eliciting multiple claims, it encourages LLMs to consider a broader range of perspectives, resulting in more balanced  responses. However, both approaches struggle to accurately simulate users. Our fine-grained analysis reveals that LLMs fail to reason over conflicting claims; and more importantly, for $50\%$ of the survey questions, the claims generated by \ours reflect a single-viewed opinion, regardless of demographic variation. Our findings reveal a fundamental limitation in LLMs’ capability of simulating user behavior: RLHF can lead to the entrenchment of certain viewpoints, therefore resistant to change across different contexts.

\begin{figure*}[t]
    \centering
    \includegraphics[width=0.32\linewidth]{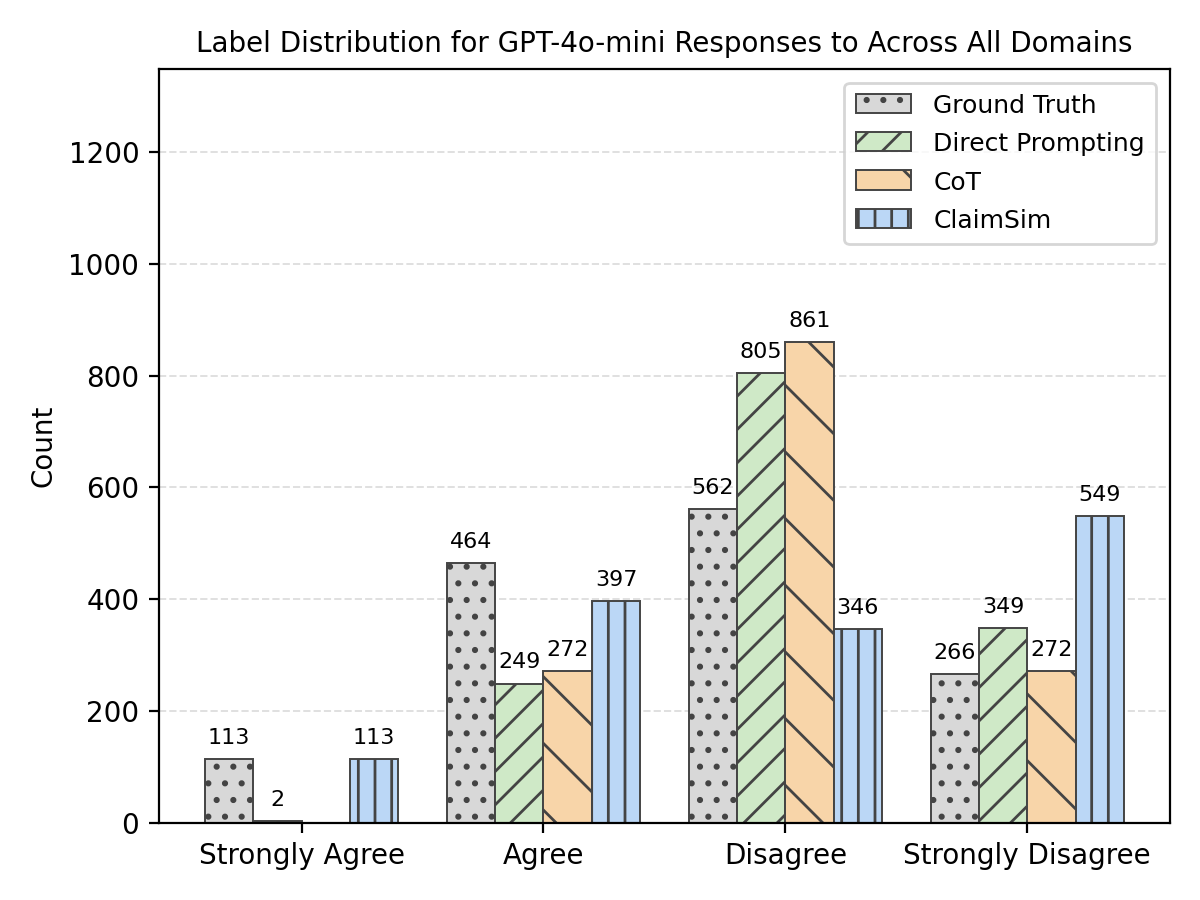}
    \hfill
    \includegraphics[width=0.32\linewidth]{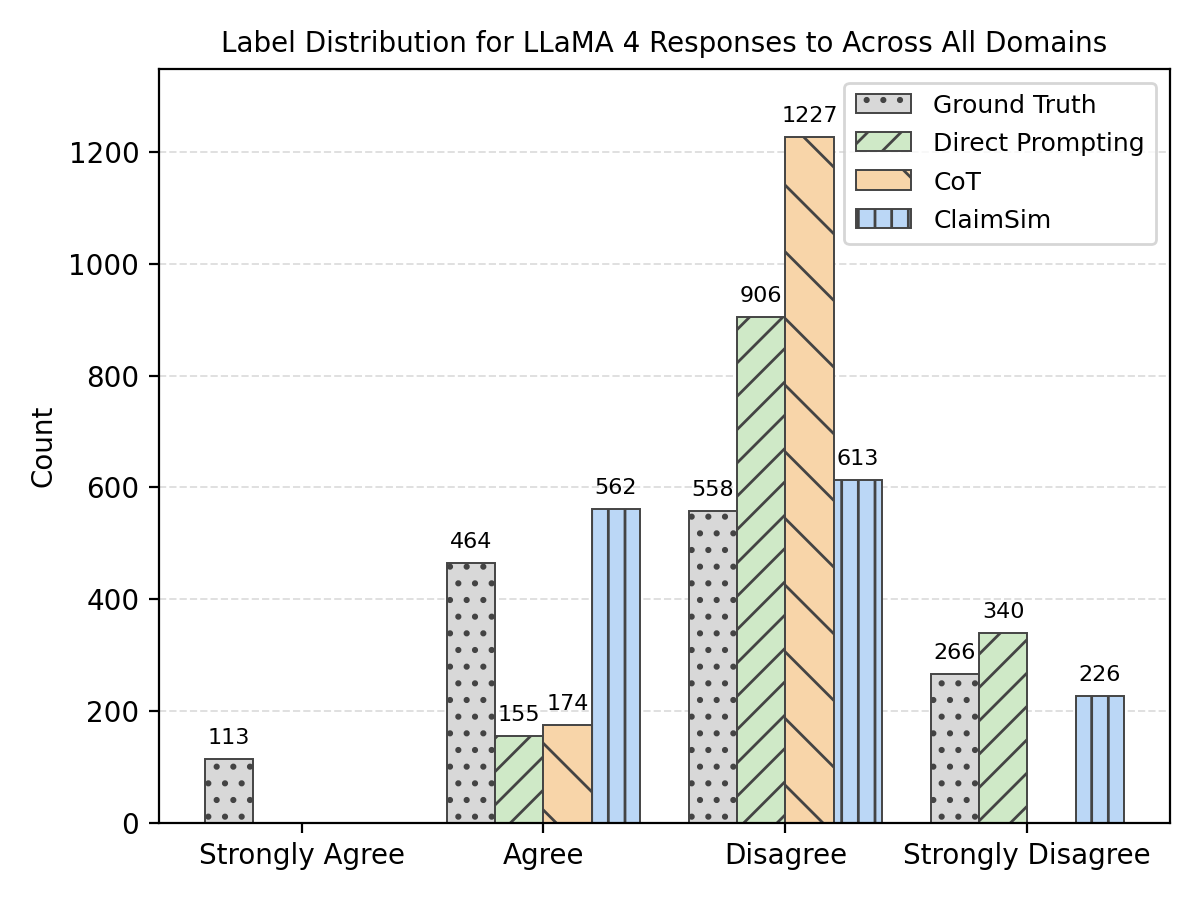}
    \hfill
    \includegraphics[width=0.32\linewidth]{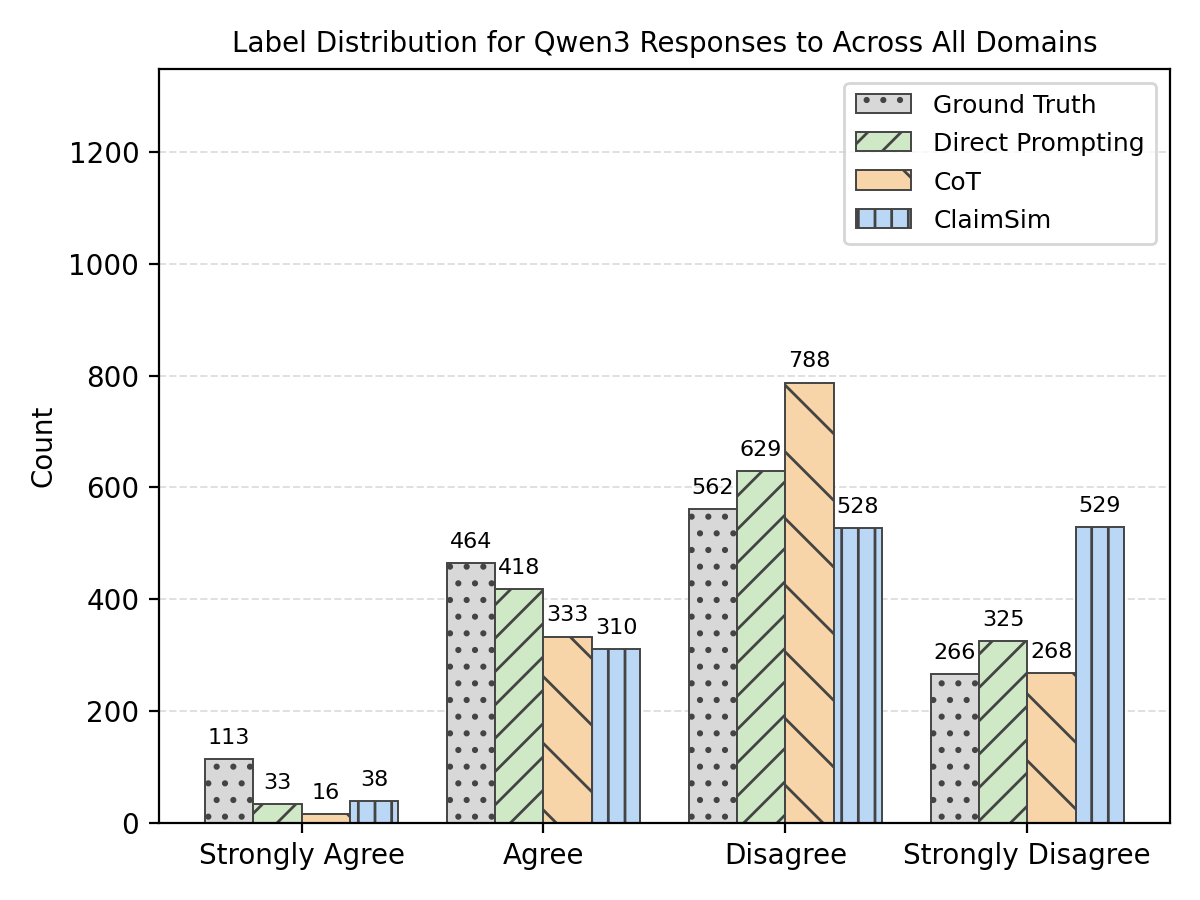}

    \caption{Comparison of answer distributions averaged across domains for \standalone, \chainot, and \ours (left to right with \textsc{GPT-4o-mini}, \textsc{Llama 4}, and \textsc{QWen 3}). \ours leads to more diverse answer distributions.}
    \label{fig:main_result}
\end{figure*}

\section{Method}
This section first introduces the task formulation of survey question answering  (\S \ref{sec:task}), followed by two approaches, chain-of-thought prompting (\S \ref{sec:baseline}), and \ours (\S \ref{sec:our-method}).

\subsection{Task Formulation}
\label{sec:task}

We formulate the task of survey question answering as follows: The input consists of a tuple of $(q,A,D)$, where $q$ refers to a multiple-choice question about opinions, \( A \) refers to the answer candidates, and \( D \) refers to a target individual's basic demographic profile, which includes attributes such as sex, birth decade, and religion, among others. The objective is to instruct an LLM to simulate the perspective of the individual with the provided demographic background:
\[
\phi_{LLM}(q, A, D) \Rightarrow a,
\]
where $a \in A$ is the most probable answer selected by an LLM.

\subsection{Direct Prompting}
\label{sec:baseline}
Our first approach uses direct prompting~\cite{beck2024sensitivity} to guide LLMs in answering survey questions. Given a question $q$ and a set of demographic features $D$, an LLM is prompted to directly generate an answer $a$.
\subsection{Chain-of-thought Prompting}
\label{sec:baseline}
We then use chain-of-thought prompting~\cite{wei2022chain} to guide an LLM to first articulate its reasoning process by considering and integrating input features, and then to generate an answer $a$ based on that reasoning step by step.

\subsection{\ours: Simulating Users with Diverse Claims Generation}
\label{sec:our-method}

We hypothesize that generating claims elicited from LLM parametric knowledge for each individual demographic feature as additional context could help mitigate model bias, as variations across features often lead to conflicting opinions. To implement this idea, we prompt LLMs several times to generate a set of representative claims $C_i = \{c_1, c_2,... c_n\}$ for each demographic feature $D_i$ based on query $q$:
\[
\phi_{LLM}(q, D_i) \Rightarrow C_i, 
\]
The LLMs are then instructed to summarize these claims $C_i$ into a single output $S_i$:
\[
\phi_{LLM}(C_i) \Rightarrow S_i,
\]
Our final answer prediction $a$ is grounded on the input query $q$, demographic information $D$, and aggregated claim summaries  $S = \{S_1, S_2,... S_k\}$:
\[
\phi_{LLM}(q, A, D,  S) \Rightarrow a.
\]


\paragraph{Diverse Claims Generation.}

For each demographic feature, we first prompt the LLM to generate claims  $C_i$ in response to the corresponding survey question $q$. To ensure diverse perspectives, we sample five separate responses, each producing one claim. We then instruct the LLM to summarize these claims into a single output $S_i$, explicitly highlighting both consistencies and contradictions within responses.




\paragraph{Answer Prediction.}
With several claim summaries derived from different demographic features, we then prompt the LLM to answer the corresponding survey question. Using chain-of-thought prompting, we guide the model to consider both the demographic information and the potentially conflicting claims, to more accurately simulate a user with a specific demographic background.

\section{Experiments}
This section first presents the dataset and experiment settings  (\S \ref{sec:dataset} \& \ref{sec:setup}). Then we provide detailed results (\S \ref{sec:results}) and analysis (\S \ref{sec:analysis}) for our survey question answering task.

\subsection{Dataset}
\label{sec:dataset}

Our dataset is derived from the World Values Survey (WVS), a comprehensive global database that captures a broad range of demographic attributes and value-based attitudes across diverse populations \cite{wvs2020}. For our experiments, we focus on three domains with the presence of strongest opinions---\textbf{gender, politics, and religion}---as a representative subset, covering a total of 16 questions. We randomly sample 100 individuals from the full dataset, ensuring substantial demographic diversity within the selection (details on sample counts are provided in the Appendix~\ref{app:sampling count}).

\subsection{Setups}
\label{sec:setup}
\paragraph{Models}
We evaluate several LLMs, including the proprietary model \textsc{GPT-4o-mini}~\cite{openai_gpt41_2025}, and open-source models \textsc{Llama4 17B}~\cite{meta_llama4_2025} and \textsc{Qwen3} 235B-A22B~\cite{qwen3_2025}. For the proprietary model, experiments are conducted via the OpenAI API. The open-source models are accessed through Together AI’s API service. For all experiments, the temperature parameter is set to 0.7 whenever applicable, as this is a common and widely used value in OpenAI’s documentation ~\cite{openai2024docs}.
\paragraph{Metrics}
Our primary evaluation metric is Answer Accuracy (\textbf{Acc}) , defined as the exact match with the gold answer option. We also report Binary Answer Accuracy (\textbf{B-Acc}), which maps nuanced labels (e.g., strongly agree and agree) into two attitudinal categories: agree and disagree. Additionally, we measure \textbf{Diversity} by analyzing the distribution of predicted answers with a histogram. To further qualitatively validate the differences between distributions produced by different methods, we compute the Wasserstein Distance, a metric that quantifies the shift between probability distributions ~\cite{villani2008optimal}.

\begin{table*}[t]
\begin{minipage}{\textwidth}
    \centering
    \small
    \setlength{\tabcolsep}{4pt}
    \resizebox{\textwidth}{!}{
    \begin{tabular}{ll|cc|cc|cc}
    \toprule
    \textbf{Base Model} & \textbf{Variant} & \multicolumn{2}{c|}{\textbf{Gender}} & \multicolumn{2}{c|}{\textbf{Politics}} & \multicolumn{2}{c}{\textbf{Religion}} \\
                        &                  & Acc & B-Acc & Acc & B-Acc & Acc & B-Acc \\
    \midrule
    \multirow{3}{*}{\textsc{GPT-4o-mini}} 
        & \standalone        & $0.40_{\scriptsize{(\pm0.01)}}$ & $\textbf{0.66}_{\scriptsize{(\pm0.02)}}$ & $\textbf{0.41}_{\scriptsize{(\pm0.05)}}$ & $\textbf{0.73}_{\scriptsize{(\pm0.01)}}$ & $0.28_{\scriptsize{(\pm0.04)}}$ & $0.52_{\scriptsize{(\pm0.03)}}$ \\
        & \textsc{CoT}      & $0.40_{\scriptsize{(\pm0.01)}}$ & $\textbf{0.66}_{\scriptsize{(\pm0.02)}}$ & $\textbf{0.41}_{\scriptsize{(\pm0.04)}}$ & $\textbf{0.73}_{\scriptsize{(\pm0.01)}}$ & $0.34_{\scriptsize{(\pm0.05)}}$ & $0.56_{\scriptsize{(\pm0.03)}}$ \\
        & \ours             & $0.30_{\scriptsize{(\pm0.04)}}$ & $\textbf{0.66}_{\scriptsize{(\pm0.02)}}$ & $\textbf{0.41}_{\scriptsize{(\pm0.05)}}$ & $0.72_{\scriptsize{(\pm0.02)}}$ & $0.32_{\scriptsize{(\pm0.04)}}$ & $0.56_{\scriptsize{(\pm0.03)}}$ \\
    \midrule
    \multirow{3}{*}{\textsc{Qwen 3}} 
        & \standalone        & $0.40_{\scriptsize{(\pm0.03)}}$ & $\textbf{0.66}_{\scriptsize{(\pm0.02)}}$ & $0.37_{\scriptsize{(\pm0.02)}}$ & $0.69_{\scriptsize{(\pm0.02)}}$ & $0.33_{\scriptsize{(\pm0.02)}}$ & $0.59_{\scriptsize{(\pm0.03)}}$ \\
        & \textsc{CoT}      & $0.41_{\scriptsize{(\pm0.01)}}$ & $\textbf{0.66}_{\scriptsize{(\pm0.01)}}$ & $0.37_{\scriptsize{(\pm0.03)}}$ & $0.68_{\scriptsize{(\pm0.02)}}$ & $0.31_{\scriptsize{(\pm0.03)}}$ & $0.59_{\scriptsize{(\pm0.02)}}$ \\
        & \ours             & $0.36_{\scriptsize{(\pm0.04)}}$ & $\textbf{0.66}_{\scriptsize{(\pm0.02)}}$ & $0.37_{\scriptsize{(\pm0.03)}}$ & $0.67_{\scriptsize{(\pm0.02)}}$ & $0.30_{\scriptsize{(\pm0.01)}}$ & $0.60_{\scriptsize{(\pm0.03)}}$ \\
    \midrule
    \multirow{3}{*}{\textsc{Llama 4}} 
        & \standalone        & $0.33_{\scriptsize{(\pm0.04)}}$ & $0.65_{\scriptsize{(\pm0.02)}}$ & $0.35_{\scriptsize{(\pm0.03)}}$ & $0.69_{\scriptsize{(\pm0.01)}}$ & $0.33_{\scriptsize{(\pm0.00)}}$ & $0.58_{\scriptsize{(\pm0.03)}}$ \\
        & \textsc{CoT}      & $\textbf{0.42}_{\scriptsize{(\pm0.02)}}$ & $0.65_{\scriptsize{(\pm0.02)}}$ & $0.32_{\scriptsize{(\pm0.03)}}$ & $0.70_{\scriptsize{(\pm0.01)}}$ & $0.31_{\scriptsize{(\pm0.02)}}$ & $0.57_{\scriptsize{(\pm0.03)}}$ \\
        & \ours             & $0.40_{\scriptsize{(\pm0.02)}}$ & $0.61_{\scriptsize{(\pm0.03)}}$ & $0.29_{\scriptsize{(\pm0.02)}}$ & $0.63_{\scriptsize{(\pm0.02)}}$ & $\textbf{0.36}_{\scriptsize{(\pm0.04)}}$ & $\textbf{0.62}_{\scriptsize{(\pm0.02)}}$ \\
    \bottomrule 
    \end{tabular}
    }
    \caption{Performance comparison between base LLMs in \standalone mode, with \chainot prompting, and with \ours built on each LLM. Results are shown as mean $\pm$ standard deviation (in smaller font). The best performance in each domain is highlighted in \textbf{bold}.}
    \label{tab:main}
\end{minipage}
\end{table*}

\begin{table}[t]
    \centering
    \normalsize 
    \setlength{\tabcolsep}{6pt} 
    \resizebox{\linewidth}{!}{ 
    \begin{tabular}{ll|c|c|c}
    \toprule
    \textbf{Base Model} & \textbf{Variant} & \textbf{Gender} & \textbf{Politics} & \textbf{Religion} \\
    \midrule
    \multirow{3}{*}{\textsc{GPT-4o-mini}} 
        & \standalone        & $0.56$ & $0.63$ & $0.70$ \\
        & \textsc{CoT}      & $0.54$ & $\textbf{0.59}$ & $0.68$ \\
        & \ours             & $\textbf{0.47}$ & $0.96$ & $\textbf{0.47}$ \\
    \midrule
    \multirow{3}{*}{\textsc{Qwen 3}} 
        & \standalone        & $0.34$ & $\textbf{0.36}$ & $0.42$ \\
        & \textsc{CoT}      & $\textbf{0.28}$ & $0.38$ & $0.45$ \\
        & \ours             & $0.59$ & $0.53$ & $\textbf{0.40}$ \\
    \midrule
    \multirow{3}{*}{\textsc{Llama 4}} 
        & \standalone        & $0.83$ & $0.76$ & $\textbf{0.53}$ \\
        & \textsc{CoT}      & $0.65$ & $0.79$ & $0.59$ \\
        & \ours             & $\textbf{0.62}$ & $\textbf{0.68}$ & $0.62$ \\
    \bottomrule
    \end{tabular}
    }
    \caption{Wasserstein distances between each method's predictions and the ground truth across different base models and domains. Lower scores indicate greater distributional similarity with the ground truth over ordinal labels. The best performance for each backbone model within each domain is highlighted in \textbf{bold}.}
    \label{tab:distance}
\end{table}

\subsection{Main Results}
\label{sec:results}
\paragraph{\standalone and \chainot do not have significant differences.}
As shown in Table~\ref{tab:main} and Table~\ref{tab:distance}, \standalone and \chainot achieve comparable Acc, B-Acc, and diversity. This suggests that a simple \chainot, lacking an explicit and verifiable reasoning path, is insufficient to effectively elicit the model’s parametric knowledge for simulating different users.

\paragraph{\ours predictions are more diverse and comparatively more align to true distributions.}
As shown in Figure~\ref{fig:main_result}, the answer distributions generated by \chainot are heavily concentrated around specific options, with this effect most pronounced in the \textsc{Llama 4} model. Table~\ref{tab:distance} further demonstrates that, in most cases, \ours achieves a closer alignment with the ground truth distribution. This suggests that \standalone and \chainot tend to reflect a unified viewpoint, even when provided with diverse demographic features. In contrast, \ours produces more diverse answer distributions by explicitly instructing LLMs to elicit multiple perspectives through diverse claim generation.


\paragraph{Nuanced difference between \ours and \chainot within the opinion category.}
As shown in Table~\ref{tab:main}, the performance gap between \chainot and \ours is notably smaller under binary labels (B-Acc) compared to finer-grained labels (Acc). This suggests that prompting LLMs to generate diverse claims primarily results in nuanced shifts within the same opinion category, e.g., from Strongly Disagree to Disagree. Fundamentally altering the model’s stance may require direct interventions, such as instruction fine-tuning or RLHF.



\paragraph{None of the approaches provide accurate user simulations.} All three methods achieve slightly above random accuracy across different backbone LLMs (Table~\ref{tab:main}), rendering the inherent difficulty of the task. Even though \ours enhances diversity and achieves better distribution alignment, these improvements are not consistently stable across different tasks (Table~\ref{tab:distance}). In the next section, we conduct a fine-grained analysis with case studies to better understand two key limitations.





\begin{figure}[!tb]
  \centering
  \begin{minipage}[t]{0.42\textwidth}
    \centering
    \includegraphics[width=1.0\textwidth]{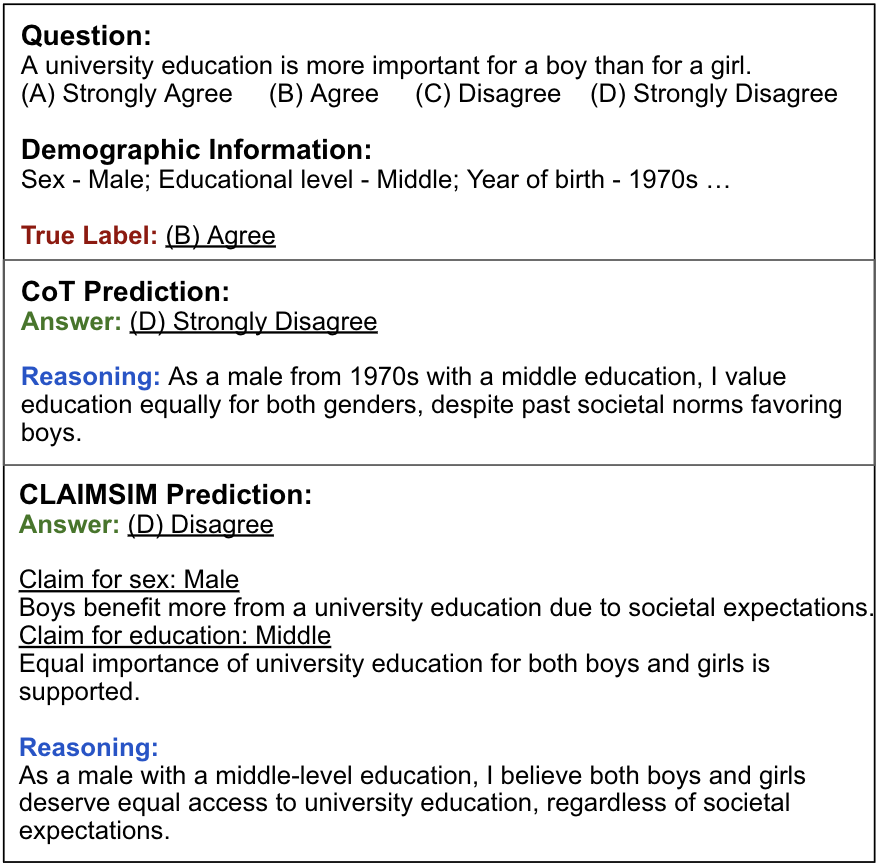}
    \caption{A case study showing LLMs fail to reason over conflicting evidence about opinions.}
    \label{fig:university}
  \end{minipage}\hfill
  \begin{minipage}[t]{0.42\textwidth}
    \centering
    \includegraphics[width=\textwidth]{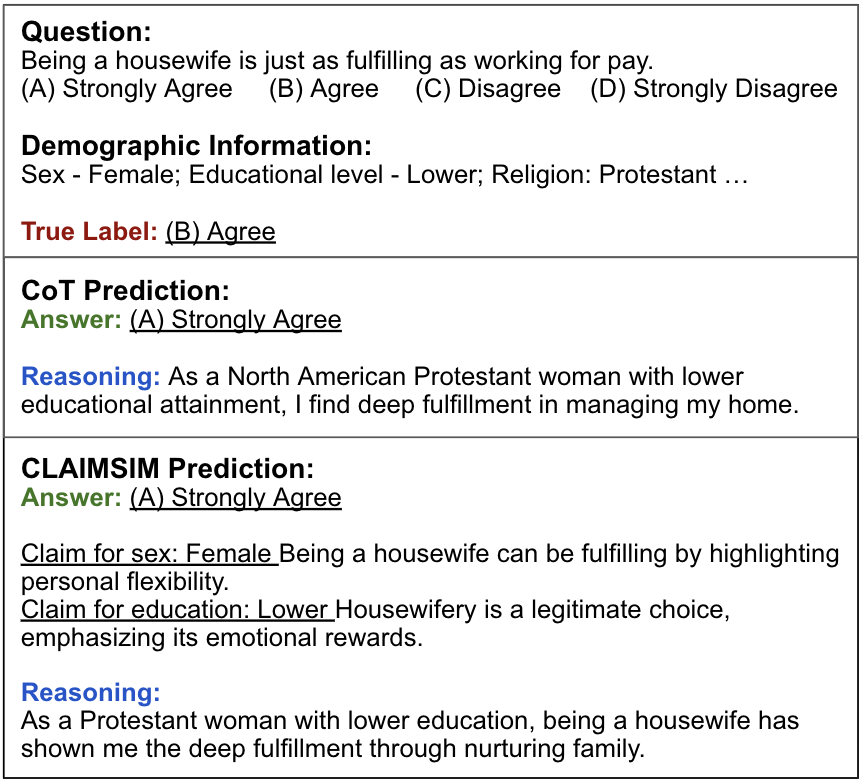}
    \caption{A case study indicating LLMs produce unified viewpoints regardless of attributes.}
    \label{fig:housewife}
  \end{minipage}
\end{figure}

\subsection{Why LLMs struggle to simulate users?}
\label{sec:analysis}

\paragraph{LLMs fail to reason over conflicting evidence about opinions.} As shown in Figure~\ref{fig:university}, although \ours elicits conflicting claims based on the \underline{sex} and \underline{education} features, LLMs fail to reason over these contradictions and generate the expected response for the target user. This remains a challenging research problem, as it requires LLMs to capture subtle relationships between demographic attributes. For example, recognizing that individuals born in the 1970s with a mid-level education may be slightly less opposed to the given view.


\paragraph{LLMs produce unified viewpoints regardless demographic attributes.} As shown in Figure~\ref{fig:housewife}, LLMs consistently generate claims that agree with the survey question, even when prompted with varying demographic profiles. Our manual analysis reveals that this pattern occurs in $50\%$ of the survey questions tested. While RLHF significantly helps align LLMs with widely accepted moral values, it also fundamentally limits their ability to simulate users whose viewpoints diverge from these norms.

\paragraph{LLMs fail to generate diverse claims for certain questions due to inherent bias.} Across five sampling rounds, LLMs successfully generate diverse claims for 60\% of questions in the gender domain, 15\% in the politics domain, and all questions in the religion domain. However, inherent biases limit diversity for some questions. For example, when asked, ``Do you think that your country's government should or should not have the right to keep people under video surveillance in public areas?'', the models consistently produce affirmative claims across all demographic groups.



\section{Related Work}

Large Language Models (LLMs) have shown strong potential in human simulation tasks~\citep{sreedhar2024simulatinghumanstrategicbehavior, binz2025centaurfoundationmodelhuman}, especially in the context of social science research involving survey questions that capture personal opinions and group-level perspectives~\cite{cao2025survey, sun2024random, kim2024aiaugmentedsurveysleveraginglarge}. For instance, \citet{cao2025survey} explore how LLMs can simulate national-level responses using supervised fine-tuning (SFT) to align model outputs with real-world distributions. At the individual level, \citet{park2023generative} propose a generative agent framework that integrates demographic profiles and interview scripts to answer survey questions. In contrast, our work focuses on investigating whether LLMs can elicit internal knowledge purely based on individual demographic attributes.

\section{Discussion}
During the elicitation of contradicting claims, we observed that the diversity of the claims are still largely constrained to LLMs' parametric knowledge. Future research shall look into effective ways for diverse claim elicitations, such as prompt optimizations~\cite{pryzant-etal-2023-automatic}.
%
We also note that answer accuracy is not the only golden metric, we found that a higher overall accuracy does not guarantee a well-aligned answer distributions. In fact, we observed \chainot and \ours are both insufficient to reflect opinions from nuanced demographic groups, highlighting the needs for further explorations on how to balance between contradicting claims and reasoning accuracy.

\section*{Limitations}
In this work, we did not study the consistency of individual-level opinions and simulation of decision-making across multiple domains, but we denote this as important perspective to realize user simulation that shall be investigated in future work.
We note that our experiments explored three representitve and recent LLMs, but did not include reasoning models such as OpenAI O3, Deepseek R1. This involves substiantially longer inference time and API cost, but may benefit both methods with more comprehensive claim generation and advanced reasoning capabilty. Also due to resource constraints, we did not explore fine-tuning or RLHF existing LLMs, despite we hypothesize that simple post-training approaches cannot tackle these limitations.
We also notice hat scaling up the domains and demographic nuances of simulated users could also bring new findings.
We leave these comparisions and experiments for future work.



\section*{Ethics Statement}
This work investigates the reliability and associated risks of large language models (LLMs) in simulating user opinions. We identify key limitations in existing approaches and demonstrate improvements in opinion diversity using our proposed method, \ours. All experiments were conducted using publicly available LLMs or APIs, and no systems were deployed in real-world settings.
Given the broader applications and societal implications of this task, we recognize several ethical concerns, including the risks of hallucinated content, overconfident claims, and the amplification of harmful biases. These issues, if left unaddressed, could lead to significant misuse or harm.
Our approach, \ours, is designed to mitigate these risks by promoting diversity and reducing bias, with the goal of supporting the development of safer and more trustworthy LLM applications in Social NLP. However, we acknowledge that limitations remain, and further investigation is necessary to fully understand and address the ethical and practical challenges posed by this work.

\section*{Acknowledgements}

 This work was supported in part through NYU Shanghai Center for Data Science, STCSM 23YF1430300 and NYU IT High Performance Computing resources, services, and staff expertise.


\bibliography{custom}

@article{sun2024random,
  title={Random silicon sampling: Simulating human sub-population opinion using a large language model based on group-level demographic information},
  author={Sun, Seungjong and Lee, Eungu and Nan, Dongyan and Zhao, Xiangying and Lee, Wonbyung and Jansen, Bernard J and Kim, Jang Hyun},
  journal={arXiv preprint arXiv:2402.18144},
  year={2024}
}

@article{cao2025survey,
  title={Specializing Large Language Models to Simulate Survey Response Distributions for Global Populations},
  author={Cao, Yong and Liu, Haijiang and Arora, Arnav and Augenstein, Isabelle and R{\"o}ttger, Paul and Hershcovich, Daniel},
  journal={arXiv preprint arXiv:2502.07068},
  year={2025}
}

@article{Santurkar2023bias2,
  title={Whose Opinions Do Language Models Reflect?},
  author={Santurkar, Shibani and Durmus, Esin and Ladhak, Faisal and Lee, Cinoo and Liang, Percy and Hashimoto, Tatsunori},
  journal={arXiv preprint arXiv:2303.17548},
  year={2023}
}

@inproceedings{wang2024not,
  title={Not All Countries Celebrate Thanksgiving: On the Cultural Dominance in Large Language Models},
  author={Wang, Wenxuan and Jiao, Wenxiang and Huang, Jingyuan and Dai, Ruyi and Huang, Jen-tse and Tu, Zhaopeng and Lyu, Michael},
  booktitle={Proceedings of the 62nd Annual Meeting of the Association for Computational Linguistics (Volume 1: Long Papers)},
  pages={6349--6384},
  year={2024}
}

@article{park2023generative,
  title={Generative Agent Simulations of 1,000 People},
  author={Joon Sung Park and Carolyn Q. Zou and Aaron Shaw and Benjamin Mako Hill and Carrie Cai and Meredith Ringel Morris and Robb Willer and Percy Liang and Michael S. Bernstein},
  journal={arXiv:2411.10109},
  year={2024},
  url={https://arxiv.org/abs/2411.10109}
}

@article{wei2022chain,
  title={Chain-of-Thought Prompting Elicits Reasoning in Large Language Models},
  author={Wei, Jason and Wang, Xuezhi and Schuurmans, Dale and Bosma, Maarten and Ichter, Brian and Xia, Fei and Chi, Ed and Le, Quoc and Zhou, Denny},
  journal={Advances in Neural Information Processing Systems},
  volume={35},
  pages={24824--24837},
  year={2022}
}

@misc{wvs2020,
  author    = {{World Values Survey}},
  title     = {World Values Survey: Wave 7 (2017-2020)},
  year      = {2020},
  url       = {https://www.worldvaluessurvey.org/wvs.jsp},
  note      = {Accessed: 2024-09-11}
}

@article{schramowski2022value,
  author = {Schramowski, Patrick and Turan, Ceyda and Andersen, Nora and Shao, Xiang and Sch{\"o}lkopf, Bernhard and Herbert, Francis and Mahendran, Arun and Hinz, Oliver and Kersting, Kristian},
  title = {Large Language Models Have General Value-Based Biases},
  journal = {Patterns},
  volume = {3},
  number = {7},
  year = {2022},
  pages = {100537},
  doi = {10.1016/j.patter.2022.100537}
}

@inproceedings{
binz2024turning,
title={Turning large language models into cognitive models},
author={Marcel Binz and Eric Schulz},
booktitle={The Twelfth International Conference on Learning Representations},
year={2024},
url={https://openreview.net/forum?id=eiC4BKypf1}
}

@misc{openai_gpt41_2025,
  title        = {Introducing GPT-4.1 in the API},
  author       = {OpenAI},
  year         = {2025},
  month        = {April},
  day          = {14},
  note         = {A new series of GPT models featuring major improvements on coding, instruction following, and long context---plus our first-ever nano model.},
  url          = {https://openai.com/index/gpt-4-1/},
  howpublished = {\url{https://openai.com/index/gpt-4-1/}}
}

@misc{meta_llama4_2025,
  title        = {The Llama 4 herd: The beginning of a new era of natively multimodal AI innovation},
  author       = {{Meta AI}},
  year         = {2025},
  month        = {April},
  day          = {5},
  url          = {https://ai.meta.com/blog/llama-4-multimodal-intelligence/},
  note         = {Official blog post introducing Llama 4 Scout, Llama 4 Maverick, and Llama 4 Behemoth models.}
}

@misc{qwen3_2025,
  title        = {Qwen3: Think Deeper, Act Faster},
  author       = {{Qwen Team}},
  year         = {2025},
  month        = {April},
  day          = {29},
  url          = {https://qwenlm.github.io/blog/qwen3/},
  note         = {Official blog post introducing Qwen3 models and their technical details.}
}

@inproceedings{pryzant-etal-2023-automatic,
    title = "Automatic Prompt Optimization with {\textquotedblleft}Gradient Descent{\textquotedblright} and Beam Search",
    author = "Pryzant, Reid  and
      Iter, Dan  and
      Li, Jerry  and
      Lee, Yin  and
      Zhu, Chenguang  and
      Zeng, Michael",
    editor = "Bouamor, Houda  and
      Pino, Juan  and
      Bali, Kalika",
    booktitle = "Proceedings of the 2023 Conference on Empirical Methods in Natural Language Processing",
    month = dec,
    year = "2023",
    address = "Singapore",
    publisher = "Association for Computational Linguistics",
    url = "https://aclanthology.org/2023.emnlp-main.494/",
    doi = "10.18653/v1/2023.emnlp-main.494",
    pages = "7957--7968"
}

@techreport{brand2024using,
  title        = {Using LLMs for Market Research},
  author       = {James Brand and Ayelet Israeli and Donald Ngwe},
  institution  = {Harvard Business School Marketing Unit},
  number       = {23-062},
  year         = {2024},
  month        = {July},
  type         = {Working Paper},
  note         = {Last revised: 29 July 2024. 46 pages.},
  url          = {https://www.hbs.edu/ris/Publication%20Files/23-062_ed720ebc-ec4d-4bc3-a6ba-bad8cfbd9d51.pdf},
  doi          = {10.2139/ssrn.4395751},
}

@inproceedings{rlhf,
author = {Ouyang, Long and Wu, Jeff and Jiang, Xu and Almeida, Diogo and Wainwright, Carroll L. and Mishkin, Pamela and Zhang, Chong and Agarwal, Sandhini and Slama, Katarina and Ray, Alex and Schulman, John and Hilton, Jacob and Kelton, Fraser and Miller, Luke and Simens, Maddie and Askell, Amanda and Welinder, Peter and Christiano, Paul and Leike, Jan and Lowe, Ryan},
title = {Training language models to follow instructions with human feedback},
year = {2022},
isbn = {9781713871088},
publisher = {Curran Associates Inc.},
address = {Red Hook, NY, USA},
booktitle = {Proceedings of the 36th International Conference on Neural Information Processing Systems},
articleno = {2011},
numpages = {15},
location = {New Orleans, LA, USA},
series = {NIPS '22}
}

@article{messeri2024artificial,
  title={Artificial intelligence and illusions of understanding in scientific research},
  author={Messeri, Lisa and Crockett, MJ},
  journal={Nature},
  volume={627},
  number={8002},
  pages={49--58},
  year={2024},
  publisher={Nature Publishing Group UK London}
}

@misc{sreedhar2024simulatinghumanstrategicbehavior,
      title={Simulating Human Strategic Behavior: Comparing Single and Multi-agent LLMs}, 
      author={Karthik Sreedhar and Lydia Chilton},
      year={2024},
      eprint={2402.08189},
      archivePrefix={arXiv},
      primaryClass={cs.HC},
      url={https://arxiv.org/abs/2402.08189}, 
}

@misc{kim2024aiaugmentedsurveysleveraginglarge,
      title={AI-Augmented Surveys: Leveraging Large Language Models and Surveys for Opinion Prediction}, 
      author={Junsol Kim and Byungkyu Lee},
      year={2024},
      eprint={2305.09620},
      archivePrefix={arXiv},
      primaryClass={cs.CL},
      url={https://arxiv.org/abs/2305.09620}, 
}

@misc{binz2025centaurfoundationmodelhuman,
      title={Centaur: a foundation model of human cognition}, 
      author={Marcel Binz and Elif Akata and Matthias Bethge and Franziska Brändle and Fred Callaway and Julian Coda-Forno and Peter Dayan and Can Demircan and Maria K. Eckstein and Noémi Éltető and Thomas L. Griffiths and Susanne Haridi and Akshay K. Jagadish and Li Ji-An and Alexander Kipnis and Sreejan Kumar and Tobias Ludwig and Marvin Mathony and Marcelo Mattar and Alireza Modirshanechi and Surabhi S. Nath and Joshua C. Peterson and Milena Rmus and Evan M. Russek and Tankred Saanum and Johannes A. Schubert and Luca M. Schulze Buschoff and Nishad Singhi and Xin Sui and Mirko Thalmann and Fabian Theis and Vuong Truong and Vishaal Udandarao and Konstantinos Voudouris and Robert Wilson and Kristin Witte and Shuchen Wu and Dirk Wulff and Huadong Xiong and Eric Schulz},
      year={2025},
      eprint={2410.20268},
      archivePrefix={arXiv},
      primaryClass={cs.LG},
      url={https://arxiv.org/abs/2410.20268}, 
}

@misc{hu2024generative,
  title={Generative language models exhibit social identity biases. arXiv [Preprint](2023)},
  author={Hu, Tiancheng and Kyrychenko, Yara and Rathje, Steve and Collier, Nigel and van der Linden, Sander and Roozenbeek, Jon},
  year={2024}
}

@article{wang2025large,
  title={Large language models that replace human participants can harmfully misportray and flatten identity groups},
  author={Wang, Angelina and Morgenstern, Jamie and Dickerson, John P},
  journal={Nature Machine Intelligence},
  pages={1--12},
  year={2025},
  publisher={Nature Publishing Group UK London}
}

@inproceedings{aher2023using,
  title={Using large language models to simulate multiple humans and replicate human subject studies},
  author={Aher, Gati V and Arriaga, Rosa I and Kalai, Adam Tauman},
  booktitle={International Conference on Machine Learning},
  pages={337--371},
  year={2023},
  organization={PMLR}
}

@inproceedings{beck2024sensitivity,
  title={Sensitivity, Performance, Robustness: Deconstructing the Effect of Sociodemographic Prompting},
  author={Beck, Tilman and Schuff, Hendrik and Lauscher, Anne and Gurevych, Iryna},
  booktitle={Proceedings of the 18th Conference of the European Chapter of the Association for Computational Linguistics (Volume 1: Long Papers)},
  pages={2589--2615},
  year={2024}
}

@misc{openai2024docs,
  author = {OpenAI},
  title = {OpenAI API Reference --- Session Object},
  howpublished = {\url{https://platform.openai.com/docs/api-reference/realtime-sessions/session_object}},
  year = {2024},
  note = {Accessed: 2025-06-30}
}

@book{villani2008optimal,
  title={Optimal transport: old and new},
  author={Villani, C{\'e}dric and others},
  volume={338},
  year={2008},
  publisher={Springer}
}
\clearpage

\appendix
\section{Sampling Count Verification}
\label{app:sampling count}
To confirm that a sample size of 100 users does not bias the experimental results, we randomly sample an additional 200 users with diverse demographic information and compare the \standalone Acc and B-Acc using a p-test (see Table~\ref{tab:sampling-acc} \& Table~\ref{tab:ptest}).

As shown in Table~\ref{tab:ptest}, all Acc and B-Acc comparisons yield p-values greater than 0.05, indicating no significant difference between sampling 100 and 200 users. These results suggest that a sample size of 100 users is sufficient and can be considered a valid choice even when scaling up.

\begin{table}[t]
    \centering
    \small
    \setlength{\tabcolsep}{4pt}
    \resizebox{\linewidth}{!}{
    \begin{tabular}{ll|cc|cc|cc}
        \toprule
        \textbf{Base Model} & \textbf{Sample Count} & 
        \multicolumn{2}{c|}{\textbf{Gender}} & 
        \multicolumn{2}{c|}{\textbf{Politics}} & 
        \multicolumn{2}{c}{\textbf{Religion}} \\
        & & Acc & B-Acc & Acc & B-Acc & Acc & B-Acc \\
        \midrule
        \multirow{2}{*}{\textsc{GPT-4o-mini}} 
            & 100 Users & 0.40 & 0.66 & 0.41 & 0.73 & 0.28 & 0.52 \\
            & 200 Users & 0.41 & 0.68 & 0.38 & 0.68 & 0.31 & 0.52 \\
        \midrule
        \multirow{2}{*}{\textsc{Qwen 3}} 
            & 100 Users & 0.40 & 0.66 & 0.37 & 0.69 & 0.33 & 0.59 \\
            & 200 Users & 0.40 & 0.67 & 0.35 & 0.70 & 0.32 & 0.59 \\
        \midrule
        \multirow{2}{*}{\textsc{Llama 4}} 
            & 100 Users & 0.33 & 0.65 & 0.35 & 0.69 & 0.33 & 0.58 \\
            & 200 Users & 0.32 & 0.63 & 0.34 & 0.67 & 0.36 & 0.57 \\
        \bottomrule
    \end{tabular}
    }
    \caption{
        Acc and B-Acc across three domains (Gender, Politics, Religion) for 100 and 200 sampled users under the \standalone setting, evaluated on three backbone models.
    }
    \label{tab:sampling-acc}
\end{table}

\begin{table}[t]
    \centering
    \small
    \setlength{\tabcolsep}{4pt}
    \resizebox{1.02\linewidth}{!}{
    \begin{tabular}{ll|cc|cc|cc}
    \toprule
    \textbf{Base Model} & \textbf{Metric} & \multicolumn{2}{c|}{\textbf{Gender}} & \multicolumn{2}{c|}{\textbf{Politics}} & \multicolumn{2}{c}{\textbf{Religion}} \\
                        &                 & t & p & t & p & t & p \\
    \midrule
    \multirow{2}{*}{\textsc{GPT-4o-mini}} 
        & Acc     & $-0.58$ & $0.591$ & $0.66$ & $0.535$ & $-2.05$ & $0.133$ \\
        & B-Acc         & $-1.09$ & $0.336$ & $1.89$ & $0.107$ & $-0.06$ & $0.958$ \\
    \midrule
    \multirow{2}{*}{\textsc{Qwen 3}} 
        & Acc     & $0.18$ & $0.867$ & $0.58$ & $0.581$ & $0.60$ & $0.592$ \\
        & B-Acc         & $-0.23$ & $0.830$ & $-0.61$ & $0.562$ & $-0.14$ & $0.898$ \\
    \midrule
    \multirow{2}{*}{\textsc{Llama 4}} 
        & Acc     & $0.27$ & $0.802$ & $1.32$ & $0.236$ & $-1.27$ & $0.294$ \\
        & B-Acc         & $0.77$ & $0.482$ & $2.28$ & $0.062$ & $0.05$ & $0.962$ \\
    \bottomrule 
    \end{tabular}
    }
    \caption{p-test comparison of Exact Accuracy and Binary Accuracy between sampling 100 users and 200 users across models and domains under \standalone setting.}
    \label{tab:ptest}
\end{table}

\section{Distributions Across Three Domains}

Figure~\ref{fig:all_distributions} presents the response distributions of three LLMs across the domains of Gender, Politics, and Religion. Each subplot illustrates how model outputs vary in opinion strength and direction within each domain.

\section{Analysis on \chainot}
As mentioned in \S \ref{sec:results}, \chainot does not improve much compared to \standalone. In this section we analyze in detail why \chainot can not improve much.

Table~\ref{tab:comparison} shows that the reasoning focus of \standalone and \chainot is quite similar, even though \chainot organizes the reasoning into several sub-points. While extensive research has demonstrated that \chainot is particularly effective for complex tasks requiring multi-step reasoning, such as solving math problems~\cite{wei2022chain}, our task \textbf{(1)} does not involve clear hierarchical reasoning steps. As a result, the outputs of \standalone and \chainot are largely comparable; \textbf{(2)} the tested LLMs struggle to autonomously construct effective reasoning processes for user simulation tasks. For example, although LLMs can break the task into steps like \textit{understanding the statement} and \textit{personal context}, they fail to fully integrate all demographic information—such as \textit{sex}—into their reasoning or determine the relative importance of each feature.

\section{Analysis on \ours}
Table~\ref{tab:claims} shows claims generated from individuals with different demographic features through five rounds of sampling. Table~\ref{tab:disparity_all} reports the percentage of pairwise demographic comparisons that exhibit disparate attitude distributions (Wasserstein distance $\geq 0.05$) across three language models (GPT, LLaMA, and QWEN) and five demographic categories (country of birth, educational level, religious denomination, sex, and year of birth) for the gender, government, and religion domains. Notably, all demographic features exhibit disparities exceeding 60\%, indicating that LLMs can reflect differences associated with diverse demographic backgrounds to an extent.

\begin{table*}[ht]
\centering
\resizebox{\textwidth}{!}{
\begin{tabular}{llccccc|c}
\toprule
\textbf{Model} & \textbf{Domain} & \textbf{Country of Birth} & \textbf{Highest Educational} & \textbf{Religious Denomination} & \textbf{Sex} & \textbf{Year of Birth} & \textbf{Row Avg} \\
\midrule
\multirow{3}{*}{\textsc{GPT-4o-MINI}} 
& Gender     & 72.0\% & 66.7\% & 58.1\% & 60.0\% & 63.8\% & \textbf{64.1\%} \\
& Government & 61.9\% & 57.1\% & 48.3\% & 42.9\% & 55.1\% & \textbf{53.1\%} \\
& Religion   & 56.7\% & 58.3\% & 57.1\% & 25.0\% & 60.7\% & \textbf{51.6\%} \\
\midrule
\multirow{3}{*}{\textsc{LLAMA 4}} 
& Gender     & 66.7\% & 73.3\% & 79.0\% & 60.0\% & 53.3\% & \textbf{66.5\%} \\
& Government & 74.0\% & 61.9\% & 76.9\% & 42.9\% & 51.7\% & \textbf{61.5\%} \\
& Religion   & 75.0\% & 75.0\% & 72.6\% & 75.0\% & 42.9\% & \textbf{68.1\%} \\
\midrule
\multirow{3}{*}{\textsc{QWEN 3}} 
& Gender     & 82.2\% & 93.3\% & 92.4\% & 60.0\% & 84.8\% & \textbf{82.5\%} \\
& Government & 74.6\% & 81.0\% & 78.2\% & 85.7\% & 70.1\% & \textbf{77.9\%} \\
& Religion   & 86.1\% & 58.3\% & 73.8\% & 50.0\% & 71.4\% & \textbf{67.9\%} \\
\midrule
\multicolumn{2}{r}{\textbf{Col Avg}} & \textbf{72.1\%} & \textbf{69.4\%} & \textbf{70.7\%} & \textbf{55.7\%} & \textbf{61.5\%} & \textbf{65.9\%} \\
\bottomrule
\end{tabular}
}

\caption{
\textbf{Wasserstein Disparity Matrix across Models and Domains.}
This table summarizes the degree of demographic disparity identified across different language models (GPT, LLAMA, and QWEN) and domains (Gender, Government, and Religion). 
Each cell reports the percentage of pairwise demographic comparisons within a given category that are considered disparate, defined as having a Wasserstein distance $\geq 0.05$. 
Higher percentages indicate greater divergence in response distributions between demographic groups. 
}
\label{tab:disparity_all}
\end{table*}

\section{Prompt Template for \standalone} 
Below we include the ~\hyperlink{standalone-prompt}{prompt template} for the \standalone method and an ~\hyperlink{standalone-example}{example prompt}.

\section{Prompt Template for \chainot} 
Below we include the ~\hyperlink{cot-template}{prompt template} for the \chainot method.

\section{Prompt Template for \ours} 
Below we include the ~\hyperlink{claim-prompt}{prompt template} for the \ours method, including the ~\hyperlink{summary-prompt}{prompt template} to generate claims for each of demographic features, generating summaries based on the claims and ~\hyperlink{simclaim-prompt}{prompt template} for generating the final responses.

\onecolumn
\begin{figure*}[t]
    \centering
    \includegraphics[width=\textwidth]{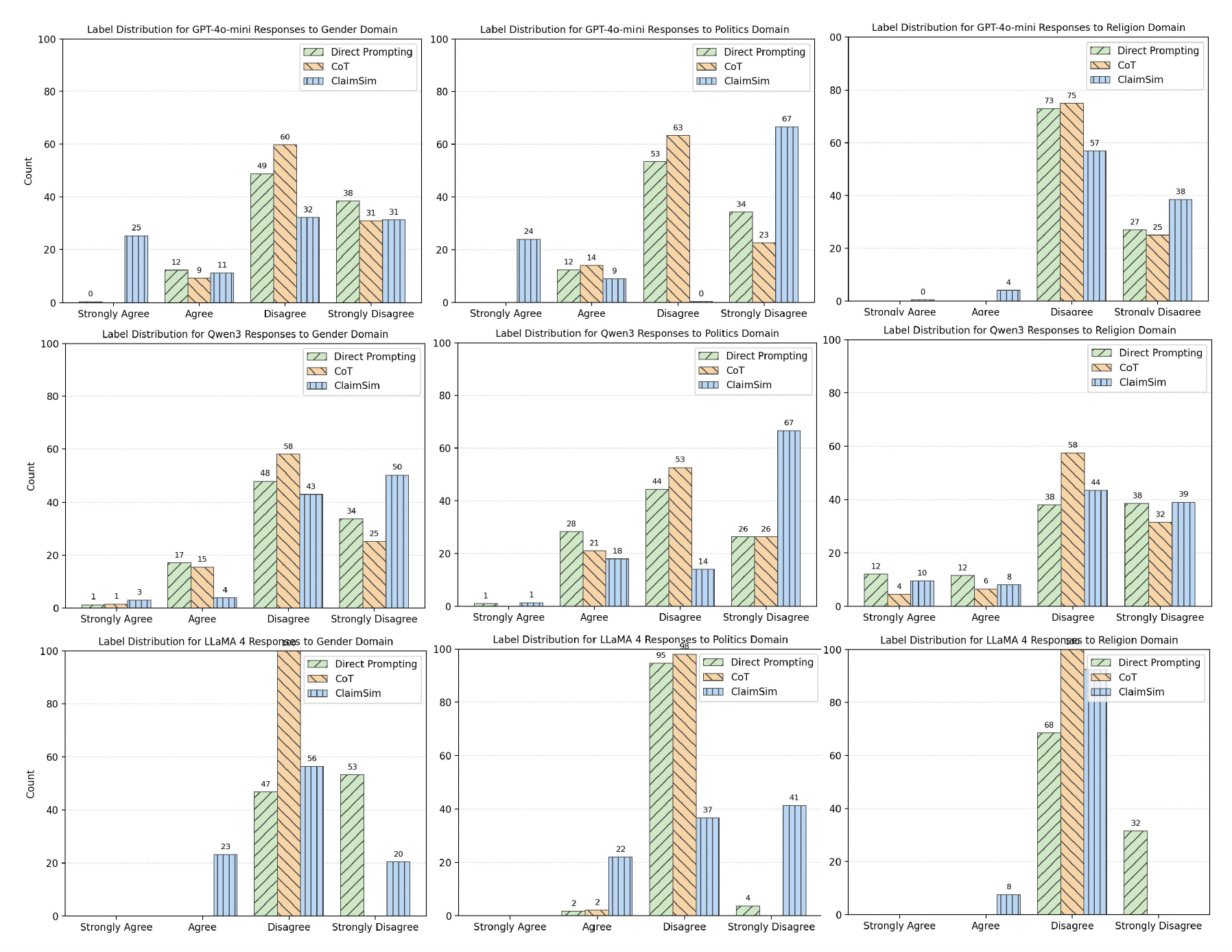}
    \caption{Response distributions of three LLMs across the Gender, Politics, and Religion domains.}
    \label{fig:all_distributions}
\end{figure*}

\begin{table}[t]
\centering
\small
\setlength{\tabcolsep}{6pt}
\begin{tabular}{p{3cm} | p{11.5cm}}
\toprule
\textbf{Category} & \textbf{Content} \\
\midrule
\textbf{Demographic Information} &
Sex: Male; Year of birth: 1950s; Age: 65 and over; Country of birth: Western Europe; Marital status: Married; Highest educational level: Higher; Occupational group: Professional/Managerial; Religious denomination: No religion \\
\midrule
\textbf{Question} &
When a mother works for pay, the children suffer. \newline
Do you strongly agree, agree, disagree, or strongly disagree? \\
\midrule
\textbf{Response (\standalone)} &
\textbf{Label:} Strongly disagree \newline
\textbf{Reasoning:} \newline
As a retired professional with higher education, I value both parents' contributions. A mother’s work brings financial stability and models independence and work ethic. The belief that children suffer is outdated; dual-income families can offer children diverse, positive experiences. \\
\midrule
\textbf{Response (\chainot)} &
\textbf{Label:} Disagree \newline
\textbf{Reasoning:} \newline
1. \textit{Understanding the statement:} Suggests maternal employment harms children—an outdated view. \newline
2. \textit{Personal context:} I grew up with working parents and value hard work and education. \newline
3. \textit{Cultural shifts:} Gender roles evolved; children can thrive in dual-income homes. \newline
4. \textit{Benefits of maternal employment:} Provides stability, models work ethic, and teaches life skills. \newline
5. \textit{Personal experience:} I’ve seen friends balance work and family successfully. \newline
6. \textit{Conclusion:} Maternal work does not inherently harm children; I disagree with the statement. \\
\midrule
\textbf{Response (\ours)} &
\textbf{Claims Summary:} \newline
1. \textit{Male:} The respondent expresses \underline{both positive and negative views}—four claims highlight benefits of maternal employment, while one suggests children suffer due to reduced care. \newline
2. \textit{Born in 1950s:} \underline{All five claims} emphasize negative impacts, focusing on reduced parental attention and emotional connection. \newline
3. \textit{Higher educational level:} All five claims support the perspective that maternal work has beneficial impacts on children, with no claims opposing this view. \newline
4. \textit{No religion:} \underline{Five claims} support this perspective, highlighting that working mothers can provide better opportunities and that the belief that children suffer from maternal employment is overly simplistic. \newline
5. \textit{Born in Western Europe:} The respondent presents a \underline{balanced view} on maternal employment, recognizing both concerns about reduced emotional support and quality time (2 claims) and the benefits of greater financial stability and opportunities for children (3 claims).\newline
\textbf{Label:} Disagree \newline
\textbf{Reasoning:} \newline
Drawing on my demographic background, I balanced the tension in the claims and weighed the benefits of maternal employment more heavily. \newline
1. \textit{Highest education level and sex:} My education encourages a nuanced view; I see both sides but value the benefits of working mothers, like financial stability and role modeling. \newline
2. \textit{Racial/ethnic group and country of birth:} Western European context supports gender equality, aligning with claims that maternal employment helps child development. \newline
3. \textit{Integration of claims:} Although reduced maternal attention is a concern, overall the positive impacts outweigh drawbacks, leading me to disagree with the statement. \\
\bottomrule
\end{tabular}
\caption{Example responses under \standalone, \chainot, and \ours modes (this example includes only a subset of personal demographic information due to space constraints). Detailed claims are shown in Table ~\ref{tab:claims}}
\label{tab:comparison}
\end{table}

\begin{table}[t]
\centering
\small
\setlength{\tabcolsep}{4pt}
\begin{tabular}{p{3cm} | p{12cm}}
\toprule
\textbf{Demographic Feature} & \textbf{Claims} \\
\midrule
\textbf{Male} &
1. When a mother works for pay, the children do not necessarily suffer; in fact, they can benefit from her employment. \newline
2. Children benefit from having a working mother. \newline
3. Children can thrive even when their mothers work for pay. \newline
4. The assertion that when a mother works for pay, the children suffer is an oversimplification that does not account for the positive impacts of maternal employment on family dynamics and child development. \newline
5. When a mother works for pay, the children suffer due to the lack of direct maternal care and attention.
\\
\midrule
\textbf{Born in 1950s} &
1. When a mother works for pay, the children suffer due to a lack of parental attention and support. \newline
2. When a mother works for pay, the children suffer in terms of emotional connection and time spent together. \newline
3. Working mothers can negatively impact their children's emotional and social development. \newline
4. When a mother works for pay, the children suffer due to reduced time and attention at home. \newline
5. When a mother works for pay, the children suffer due to reduced maternal presence and support.
\\
\midrule
\textbf{Higher Educational Level} &
1. The assertion that when a mother works for pay, the children suffer is an oversimplification that ignores the positive impacts of maternal employment on both the mother and her children. \newline
2. When a mother works for pay, the children can benefit from increased financial stability and role modeling of work ethics. \newline
3. A mother's paid work does not inherently harm children; in fact, it can provide them with better opportunities. \newline
4. Working mothers can provide better opportunities for their children. \newline
5. A mother’s work for pay can positively influence her children’s development and wellbeing.
\\
\midrule
\textbf{No Religion} &
1. A mother's decision to work for pay does not necessarily mean that her children suffer; in fact, it can have positive effects on their development and wellbeing. \newline
2. Working mothers can provide better opportunities for their children. \newline
3. The belief that children suffer when their mothers work for pay is overly simplistic and does not account for the diverse realities of modern families. \newline
4. Working mothers can positively influence their children's development and wellbeing. \newline
5. Children can thrive even when their mothers work for pay.
\\
\midrule
\textbf{Born in Western Europe} &
1. The claim that when a mother works for pay, the children suffer is overly simplistic and does not consider the multifaceted benefits of maternal employment. \newline
2. Mothers working for pay can provide better financial stability and opportunities for their children. \newline
3. Working mothers can provide better opportunities for their children. \newline
4. When a mother works for pay, the children may suffer in terms of emotional support and quality time spent together. \newline
5. When a mother works for pay, it can lead to negative consequences for children, including reduced quality time and emotional support.
\\
\bottomrule
\end{tabular}
\caption{Example claims for each of five demographic features in response of the survey question "Do you agree that when a mother works for pay, the children suffer."}
\label{tab:claims}
\end{table}

\clearpage
\begin{tcolorbox}[
    colback=white, 
    colframe=black, 
    title={Prompt Template for \standalone}, 
    label={box:standalone-prompt}, 
    width=\textwidth,      
    boxsep=4pt,            
    left=4pt,
    right=4pt
    ]
\textcolor{myred}{\textbf{\# Personal Demographic Information}}\\
The respondent's demographic details are as follows:\\
\{\texttt{\% for category, value in demo\_infos.items() \%}\}\\
\textcolor{mygreen}{\texttt{- \{\{ category \}\}: \{\{ value \}\}}}\\
\{\texttt{\% endfor \%}\}\\

\textcolor{myred}{\textbf{\# Task}}\\
Imagine you are the respondent. Based on your demographic background, answer the following question under the topic of \textcolor{mygreen}{\texttt{\{\{ domain \}\}}}.\\

\textcolor{myred}{\textbf{\# Instruction}}\\
\textcolor{mygreen}{\texttt{\{\{ instruction \}\}}}\\

\textcolor{myred}{\textbf{\# Question}}\\
\textcolor{mygreen}{\texttt{\{\{ question \}\}}}\\

\textcolor{myred}{\textbf{\# Label Choices}}\\
You must choose \textbf{exactly one} label from the options below:\\
\textcolor{mygreen}{\texttt{\{\{ labels \}\}}}\\

\textcolor{myred}{\textbf{\# Response Format}}\\
-Label: The selected label from the provided choices.\\
-Reasoning: An explanation leading to your choice.
\end{tcolorbox}

\clearpage
\begin{tcolorbox}[
    colback=white, 
    colframe=black, 
    title={Prompt Example for \standalone}, 
    label={box:standalone-example}, 
    width=\columnwidth,     
    boxsep=4pt,             
    left=4pt,
    right=4pt,
    sharp corners
    ]
\textcolor{myred}{\textbf{\# Personal Demographic Information}}\\
The respondent's demographic details are as follows:\\

\textcolor{myred}{-} \textbf{Sex}: Male\\
\textcolor{myred}{-} \textbf{Year of birth}: 1940s\\
\textcolor{myred}{-} \textbf{Age}: 65 and over\\
\textcolor{myred}{-} \textbf{Respondent immigrant}: I am an immigrant to this country (born outside this country)\\
\textcolor{myred}{-} \textbf{Mother immigrant}: Immigrant\\
\textcolor{myred}{-} \textbf{Father immigrant}: Immigrant\\
\textcolor{myred}{-} \textbf{Country of birth: Respondent}: East Asia\\
\textcolor{myred}{-} \textbf{Respondent citizen}: yes\\
\textcolor{myred}{-} \textbf{Number of people in household}: 1\\
\textcolor{myred}{-} \textbf{Do you live with your parents}: Not live with parents\\
\textcolor{myred}{-} \textbf{Marital status}: Married\\
\textcolor{myred}{-} \textbf{How many children do you have}: 1 child\\
\textcolor{myred}{-} \textbf{Highest educational level: Respondent}: Higher\\
\textcolor{myred}{-} \textbf{Employment status}: Retired/pensioned\\
\textcolor{myred}{-} \textbf{Employment status - Respondent´s Spouse}: Full time (30 hours a week or more)\\
\textcolor{myred}{-} \textbf{Respondent - Occupational group}: Professional/Managerial\\
\textcolor{myred}{-} \textbf{Sector of employment}: Private business or industry\\
\textcolor{myred}{-} \textbf{Are you the chief wage earner in your house}: Yes\\
\textcolor{myred}{-} \textbf{Family savings during past year}: Save money\\
\textcolor{myred}{-} \textbf{Social class}: Lower class\\
\textcolor{myred}{-} \textbf{Income level}: Low\\
\textcolor{myred}{-} \textbf{Religious denominations - major groups}: Protestant\\

\vspace{1em}

\textcolor{myred}{\textbf{\# Task}}\\
Imagine you are the respondent. Based on your demographic background, answer the following question under the topic of \textbf{gender}.\\

\textcolor{myred}{\textbf{\# Instruction}}\\
For each of the following statements I read out, can you tell me how strongly you agree or disagree with each. Do you strongly agree, agree, disagree, or strongly disagree?\\

\textcolor{myred}{\textbf{\# Question}}\\
When a mother works for pay, the children suffer.\\

\textcolor{myred}{\textbf{\# Label Choices}}\\
You must choose \textbf{exactly one} label from the options below:\\
\textcolor{mygreen}{['strongly agree', 'agree', 'disagree', 'strongly disagree']}\\

\textcolor{myred}{\textbf{\# Response Format}}\\
-Label: The selected label from the provided choices.\\
-Reasoning: An explanation leading to your choice.
\end{tcolorbox}

\onecolumn
\begin{tcolorbox}[
    colback=white, 
    colframe=black, 
    title={Prompt Template for \chainot}, 
    label={box:cot-template}, 
    width=\textwidth,   
    boxsep=4pt,         
    left=4pt,
    right=4pt,
    sharp corners
    ]
\textcolor{myred}{\textbf{\# Personal Demographic Information}}\\
The respondent's demographic details are as follows:\\
\{\texttt{\% for category, value in demo\_infos.items() \%}\}\\
\textcolor{mygreen}{\texttt{- \{\{ category \}\}: \{\{ value \}\}}}\\
\{\texttt{\% endfor \%}\}\\

\textcolor{myred}{\textbf{\# Task}}\\
Imagine you are the respondent. Based on your demographic background, thoughtfully answer the following question under the topic of \textcolor{mygreen}{\texttt{\{\{ domain \}\}}}.\\
Before selecting your final answer, you must \textbf{reason step-by-step} to demonstrate your thought process.\\

\textcolor{myred}{\textbf{\# Instruction}}\\
\textcolor{mygreen}{\texttt{\{\{ instruction \}\}}}\\

\textcolor{myred}{\textbf{\# Question}}\\
\textcolor{mygreen}{\texttt{\{\{ question \}\}}}\\

\textcolor{myred}{\textbf{\# Label Choices}}\\
You must choose \textbf{exactly one} label from the options below:\\
\textcolor{mygreen}{\texttt{\{\{ labels \}\}}}\\

\textcolor{myred}{\textbf{\# Response Format}}\\
- Label: The selected label from the provided choices.\\
- Reasoning: A detailed step-by-step explanation leading to your choice, following a chain-of-thought approach.\\
Important: If you do not provide a detailed reasoning step-by-step, your answer will be considered incomplete.
\end{tcolorbox}
\twocolumn

\onecolumn
\begin{tcolorbox}[
    colback=white, 
    colframe=black, 
    title={Claim Generation Prompt}, 
    label={box:claim-prompt}, 
    width=\textwidth,       
    boxsep=4pt,             
    left=4pt,
    right=4pt,
    sharp corners
    ]
\textcolor{myred}{\textbf{\# Instruction}}\\
Suppose you are a respondent with the following demographic information:\\
- \textcolor{mygreen}{\texttt{\{\{ feature\_category \}\}}}: \textcolor{mygreen}{\texttt{\{\{ feature\_label \}\}}}\\

\textcolor{myred}{\textbf{\# Related Question}}\\
- Question: \textcolor{mygreen}{\texttt{\{\{ question \}\}}}\\

\textcolor{myred}{\textbf{\# Task}}\\
Please provide all the viewpoints you can think of under the related question. Remember to ground every single opinion in \textbf{your} specific background and lived experience. Take reality and ideals into account. For each opinion, include: \\

- Topic (e.g., the keyword of the opinion)\\
- Claim (A clear expression of the viewpoint)\\
- Evidence\_for\_claim (e.g., the evidence to support the claim)\\
- Counterpoint (the opposite of the viewpoint)\\
- Evidence\_for\_counterpoint (e.g., the evidence to support the counterpoint)\\

List \textbf{1} most representative opinion in plain text.\\

\textcolor{myred}{\textbf{\# Output Format}} (Required)\\
- \textbf{Claim}: Your claim for this question.
\end{tcolorbox}
\twocolumn

\onecolumn
\begin{tcolorbox}[
    colback=white, 
    colframe=black, 
    title={Prompt for Summary Generation}, 
    label={box:summary-prompt}, 
    width=\textwidth,      
    boxsep=4pt,            
    left=4pt,
    right=4pt,
    sharp corners
    ]
\textcolor{myred}{\textbf{\# Task}}\\
Below are some claims from the respondent with the following demographic feature:\\
- \textcolor{mygreen}{\texttt{\{\{ feature\_category \}\}}}: \textcolor{mygreen}{\texttt{\{\{ feature\_label \}\}}}\\

Please provide a concise summary that captures the key perspectives expressed in the claims.\\

\textcolor{myred}{\textbf{\# Claims}}\\
\{\texttt{\% for claim in claims \%}\}\\
- \textbf{Claim}: \textcolor{mygreen}{\texttt{\{\{ claim \}\}}}\\
\{\texttt{\% endfor \%}\}\\

\textcolor{myred}{\textbf{\# Output Format (Required)}}\\
- Summary: Provide a 2-3 sentence synthesis of the respondent's views, clearly identifying key themes, contradictions, or tensions. Explicitly state how many claims support one perspective versus how many support an opposing or contrasting view, if applicable.
\end{tcolorbox}

\onecolumn
\begin{tcolorbox}[
    colback=white, 
    colframe=black, 
    title={Prompt Template for \ours}, 
    label={box:simclaim-prompt}, 
    width=\textwidth,      
    boxsep=4pt,            
    left=4pt,
    right=4pt,
    sharp corners
    ]
\textcolor{myred}{\textbf{\# Demographic Profile}}\\
Below is a simulated demographic profile. Please respond as if you belong to this background:\\
\{\texttt{\% for category, value in demo\_infos.items() \%}\}\\
\texttt{- \textcolor{mygreen}{\{\{ category \}\}}: \textcolor{mygreen}{\{\{ value \}\}}}\\
\{\texttt{\% endfor \%}\}\\

\textcolor{myred}{\textbf{\# Context}}\\
You will evaluate claims and counterpoints that reflect opinions or beliefs people might hold in the field of \textcolor{mygreen}{\texttt{\{\{ domain \}\}}}. Each pair is designed to capture a possible tension or debate that may arise based on demographic perspectives.\\

These statements are \textit{hypothetical and intentionally diverse} to explore how views might vary across backgrounds. Your task is not to judge them by factual accuracy, but to engage thoughtfully based on your assigned profile.\\

\textcolor{myred}{\textbf{\# Statements}}\\
For each of the following, consider both the claim and the counterpoint:\\
\{\texttt{\% for claim, counterpoint in claims.items() \%}\}\\
- \textbf{Claim}: \textcolor{mygreen}{\texttt{\{\{ claim \}\}}}\\
- \textbf{Counterpoint}: \textcolor{mygreen}{\texttt{\{\{ counterpoint \}\}}}\\
\{\texttt{\% endfor \%}\}\\

\textcolor{myred}{\textbf{\# Task}}\\
You are asked to select the position that would most closely align with the simulated demographic perspective above. This is a reasoned choice based on how someone from this profile might respond. In doing so, carefully consider how the specific claims made in the prompt may influence their reasoning. At the same time, critically reflect on potential counterpoints---how someone from this demographic might still be persuaded by alternative views. Your answer should weigh these tensions and offer a thoughtful justification.\\

\textcolor{myred}{\textbf{\# Instruction}}\\
\textcolor{mygreen}{\texttt{\{\{ instruction \}\}}}\\

\textcolor{myred}{\textbf{\# Question}}\\
\textcolor{mygreen}{\texttt{\{\{ question \}\}}}\\

\textcolor{myred}{\textbf{\# Label Choices}}\\
Choose \textbf{exactly one} of the following:\\
\textcolor{mygreen}{\texttt{\{\{ labels \}\}}}\\

\textcolor{myred}{\textbf{\# Response Format (Required)}}\\
- \textbf{Label}: your selected label from above\\
- \textbf{Reasoning}: Step-by-step explanation of how the claims, counterpoints and simulated demographic background influence the choice. Be specific and avoid generic justifications.\\

\textbf{> Warning: Incomplete responses without detailed reasoning will be considered invalid for this task.}
\end{tcolorbox}
\twocolumn


\end{document}